\documentclass[conference]{IEEEtran}
\IEEEoverridecommandlockouts

\usepackage{amsmath,amssymb,amsfonts}
\usepackage{algorithmic}
\usepackage{graphicx}
\usepackage{textcomp}
\usepackage{xcolor}
\usepackage{placeins}
\usepackage{times}
\usepackage[T5]{fontenc}
\usepackage[utf8]{inputenc}
\usepackage{microtype}
\usepackage{multirow}
\usepackage{graphicx}
\usepackage{marvosym}
\usepackage{float}
\usepackage{amsmath,amssymb,amsfonts}
\usepackage{pifont}
\usepackage{siunitx}
\usepackage{latexsym}
\usepackage{etoolbox}
\usepackage{balance}

\usepackage[numbers]{natbib}
\usepackage[bookmarks,bookmarksopen,bookmarksdepth=3,bookmarksnumbered]{hyperref}
\hypersetup{colorlinks, citecolor=blue, linkcolor=blue, urlcolor=blue}

\usepackage[left=2.5cm,right=2.5cm,top=1.75cm,bottom=1.5cm]{geometry}

\PassOptionsToPackage{bookmarks=false}{hyperref}

\begin{document}

\title{Is word segmentation necessary for\\ Vietnamese sentiment classification?}

\author{\IEEEauthorblockN{\textbf{Duc-Vu Nguyen$^{\text{1, 2}}$, Ngan Luu-Thuy Nguyen$^{\text{1, 2}}$}}
\IEEEauthorblockA{$^{\text{1}}$University of Information Technology, Ho Chi Minh City, Vietnam \\
$^{\text{2}}$Vietnam National University, Ho Chi Minh City, Vietnam\\
\texttt{\{vund,ngannlt\}@uit.edu.vn}}
}

\maketitle

\begin{abstract}
To the best of our knowledge, this paper made the first attempt to answer whether word segmentation is necessary for Vietnamese sentiment classification. To do this, we presented five pre-trained monolingual S4-based language models for Vietnamese, including one model without word segmentation, and four models using \texttt{RDRsegmenter}, \texttt{uitnlp}, \texttt{pyvi}, or \texttt{underthesea} toolkits in the pre-processing data phase. According to comprehensive experimental results on two corpora, including the VLSP2016-SA corpus of technical article reviews from the news and social media and the UIT-VSFC corpus of the educational survey, we have two suggestions. Firstly, using traditional classifiers like Naive Bayes or Support Vector Machines, word segmentation maybe not be necessary for the Vietnamese sentiment classification corpus, which comes from the social domain. Secondly, word segmentation is necessary for Vietnamese sentiment classification when word segmentation is used before using the BPE method and feeding into the deep learning model. In this way, the \texttt{RDRsegmenter} is the stable toolkit for word segmentation among the \texttt{uitnlp}, \texttt{pyvi}, and \texttt{underthesea} toolkits.
\end{abstract}


\section{Introduction}
\label{sec:introduction}
Word segmentation is a fundamental problem in the field of Vietnamese natural language processing. For instance, the Vietnamese text ``hiện đại hóa đất nước'' ($\text{modernize}_\text{hiện\_đại\_hoá}$ $\text{country}_\text{đất\_nước}$), which consists of five syllables, is segmented into ``hiện\_đại\_hoá đất\_nước''. Underscores indicate the white spaces function as syllable separators, and white spaces are used for word splits. That is a challenging problem in the early stage of natural language processing in Vietnamese \cite{diend1}. Common tasks in Vietnamese syntax analysis tasks, such as part-of-speech tagging \cite{lehongtag10}, constituency parsing \cite{nguyen-etal-2009-building, 10.1007/s10579-017-9398-3}, dependency parsing \cite{10.1007/978-3-319-07983-7_26,6719884,nguyen-2018-bktreebank}, and semantic parsing \cite{JCSCE}, are required to undergo word segmentation. In these tasks, the mistake of word segmentation directly affects them. Therefore, many previous works introduced various approaches to improve Vietnamese word segmentation performance, including single word segmentation task only \cite{diend1, dinhdien06, ha2003, nguyenetal2006, thangdq1, hongphuong08, songnguyen16, phongnt1, datnq1, vund2018, spanviws} and multi-task containing word segmentation and part-of-speech tagging \cite{ducpd1, nguyen-etal-2017-word}, dependency parsing \cite{nguyen-2019-neural}. From the previous works above, word segmentation is obligatory for Vietnamese syntax analysis tasks, including part-of-speech tagging, constituency parsing, dependency parsing, and semantic parsing.

In addition to the Vietnamese syntax analysis tasks, sentiment classification attracts many researchers because we can apply it quickly to real-life applications. Firstly, \citet{5632131} introduced a system that can classify a computer product review into positive or negative. The study of \citet{5632131} is the first work that analyzes sentiment at the sentence level in Vietnamese. Another example, in the research of \citet{7043403}, they proposed a framework classifying the hotel review into one positive, neutral, or negative class. Almost all do word segmentation in data pre-processing before researching methodology to solve the sentiment classification task. The \texttt{vnTokenizer} toolkit \cite{hongphuong08} was used for pre-processing the VLSP2016-SA \cite{vlsp2016sa} and VS corpora in the work of \citet{8119429}, the hotel reviews corpus proposed by \citet{7043403}, the electronic devices corpus proposed by \citet{7371776}, the education survey corpus proposed by \citet{bagofstruct}, and the online reviews corpus proposed by \citet{Tran2017}. Another Vietnamese word segmenter proposed by \citet{ducpd1}, was used in the works of \citet{10.1007/978-3-642-23620-4_21}, \citet{5632131}, and \citet{Vu2011} about product reviews domain. After that, \citet{10.1145/2676585.2676606} used the \texttt{JVnSegmenter} toolkit \cite{nguyenetal2006} for pre-processing their food reviews corpus. In addition, \citet{7758052} apply the \texttt{UETsegmenter} \cite{phongnt1} to preproces their mobile product reviews corpus. Lastly, the works \cite{8573351, 9335912} about sentiment analysis on feedback of students \cite{vsfc} and \cite{10.1007/978-981-15-6168-9_27} about emotion recognition used the \texttt{RDRsegmenter} of \citet{datnq1} for their research. All of the above works on Vietnamese sentiment analysis used Vietnamese word segmenter published in scientific papers. Moreover, we can survey two well-known Vietnamese word segmenters, namely \texttt{pyvi}\footnote{\url{https://pypi.org/project/pyvi/}} and \texttt{underthesea}\footnote{\url{https://pypi.org/project/underthesea/}} scientifically unpublished up to now. For instances, the \texttt{pyvi} toolkit was used in the research of \citet{10.1007/978-981-15-6168-9_15} about sentiment analysis on VLSP2018-SA corpus \cite{vlsp2016sa} and research of \citet{DBLP:conf/somet/0002LTLHP21} on product reviews. Another instance, \citet{DBLP:conf/enase/0002HHPZ20} used \texttt{underthesea} toolkit for pre-processing their electronic products comments dataset. Lastly, we observed many prior studies on Vietnamese sentiment classification apply word segmentation in the pre-processing phase.

Although we can observe that the word segmentation phase is almost necessary for the Vietnamese sentiment classification problem, recently, there have been some studies on the Vietnamese sentiment classification problem without the word segmentation phase. For example, the research \cite{10.1007/978-3-030-82147-0_53} applied the fastText \cite{joulin2017bag} model for pre-processing and embedding the input data without the Vietnamese word segmentation phase. For other examples, the research \cite{9140757, 9335899, 9287650} studied the sentiment classification problem using the pre-trained multilingual language model mBERT \cite{devlin-etal-2019-bert}, which is not required Vietnamese word segmentation. To our best knowledge, the studies \cite{10.1007/978-3-030-82147-0_53,9140757, 9335899, 9287650} applied sub-words \cite{bojanowski2017enriching} method for the better handling unseen words. On the other hand, the fastText and mBERT models are made for multilingual purposes. Hence, the studies \cite{10.1007/978-3-030-82147-0_53,9140757, 9335899, 9287650} did not use the word segmentation phase for the Vietnamese sentiment classification problem. Indeed, the research \cite{9335912} used \texttt{RDRsegmenter} toolkit for data pre-processing before using the pre-trained monolingual PhoBERT model \cite{phobert}, which is made for Vietnamese and applied Byte-Pair Encoding (BPE) method \cite{sennrich-etal-2016-neural} for sub-word representations for Vietnamese.

From the above observations, we have to admit that word segmentation is crucial for Vietnamese syntax analysis tasks, including part-of-speech tagging, constituency parsing, dependency parsing, and semantic parsing. That strongly motivates many proposed Vietnamese word segmentation methods in prior studies. On the Vietnamese Treebank corpus for word segmentation \cite{nguyen-etal-2009-building}, the highest F-score is 98.31\% achieved by the span labeling approach \cite{spanviws} using XLM-RoBERTa \cite{conneau-etal-2020-unsupervised}, which is very slow when inference on CPU device. However, the well-known Vietnamese word segmenter toolkit \texttt{RDRsegmenter} achieved the F-score of 97.90\% by the rule approach with a fast speed for inference. In addition, from the above observations, word segmentation is used widely in many prior studies on Vietnamese sentiment classification, while some studies did not use word segmentation. Consequently, we have raised a research question, ``\textit{\textbf{Is word segmentation necessary for Vietnamese sentiment classification?}}''

To attempt to answer the question above, we used four Vietnamese word segmentation toolkits\footnote{Because of time limitations, we only selected the most recently published Vietnamese word segmentation toolkits. We will expand our research on other toolkits in future work.}. Firstly, we chose the fast and accurate Vietnamese word segmentation toolkit \texttt{RDRsegmenter} \cite{datnq1}. Secondly, we chose the Vietnamese word segmentation toolkit \texttt{uitnlp} \cite{vund2018}, which is proposed for ambiguity reduction and suffix capture. Lastly, we chose two well-known Vietnamese word segmentation toolkits, including \texttt{pyvi} and \texttt{underthesea} scientifically unpublished up to now. Regarding corpora, we chose two related to the Vietnamese sentiment classification problem, including the VLSP2016-SA corpus \cite{vlsp2016sa} of technical article reviews from the news and social media and the UIT-VSFC corpus of educational survey \cite{vsfc}. Regarding classifiers, we chose two traditional classifiers as baselines, including Naive Bayes (NB) and Support Vector Machines (SVMs) and the recent modern text encoder, namely, the Structured State Space Sequence model (S4) \cite{s4}.

In summary our contributions are the following:
\begin{itemize}
    \item Five pre-trained monolingual S4-based language models for Vietnamese, including one model without word segmentation, and four models using \texttt{RDRsegmenter} \cite{datnq1}, \texttt{uitnlp} \cite{vund2018}, \texttt{pyvi}, or \texttt{underthesea} toolkits in the pre-processing data phase.
    \item According to extensive experimental results on two corpora, including the VLSP2016-SA corpus of technical article reviews from the news and social media and the UIT-VSFC corpus of the educational survey, we have two suggestions:
    \begin{itemize}
        \item Using traditional classifiers like Naive Bayes or Support Vector Machines, word segmentation maybe not be necessary for the Vietnamese sentiment classification corpus, which comes from the social domain.
        \item Word segmentation is necessary for Vietnamese sentiment classification when word segmentation is considered pre-processing before using the BPE method and feeding into the deep learning model. By this way, the \texttt{RDRsegmenter} is the stable toolkit for word segmentation among the \texttt{uitnlp}, \texttt{pyvi}, and \texttt{underthesea} toolkits.
    \end{itemize}
\end{itemize}

\section{Pre-Trained Monolingual S4-Based Language Models For Vietnamese}
\label{sec:pre-trained}

In this paper, we focus on finding the effect of different word segmentation toolkits in the pre-processing data phase on deep learning models' Vietnamese sentiment classification performances. Hence, this section describes the models that applied different word segmentation toolkits, used the same S4-based language model architecture, and trained on the same pre-training data with the same optimization algorithm.

\subsection{Architecture}
\label{subsec:architecture}
This year, \citet{s4} suggested that the Structured State Space Sequence model (S4) has the potential to be an effective general sequence modeling solution. Hence, we chose S4 model \cite{s4} to develop pre-trained monolingual S4-based language models for Vietnamese. Besides, we have to train up to five monolingual S4-based language models for Vietnamese, including one model without word segmentation, and four models using \texttt{RDRsegmenter}, \texttt{uitnlp}, \texttt{pyvi}, or \texttt{underthesea} toolkits in the pre-processing data phase. Therefore, we used only two blocks of S4 layers (instead of 16 blocks in the original study \cite{s4}) alternated with position-wise feedforward layers, with a feature dimension of 256. Following \cite{s4}, we used a GLU activation after the S4 linear layer and used two S4 layers per block. The embedding and softmax layers were the Adaptive Embedding from \cite{DBLP:journals/corr/abs-1809-10853} with customized cutoffs 5000, 10000, 15000. The embedding size is 256, and the vocabulary size is 64K, as we will in the following subsection (\ref{subsec:pre-training-data}). Therefore, there are about 20M parameters for each such model, smaller than about six times compared with $\text{PhoBERT}_\text{base}$ \cite{phobert}.

\subsection{Pre-training Data}
\label{subsec:pre-training-data}
We used a 10 GB pre-training dataset, is the concatenation of two corpora, the first one is the Vietnamese Wikipedia corpus\footnote{\url{https://github.com/NTT123/viwik18/tree/viwik19}} ($\sim$ 1 GB), and the second corpus $(\sim$ 9 GB) is apart from the 18.6 GB Vietnamese news corpus\footnote{\url{https://github.com/binhvq/news-corpus\#full-txttitle--description--body-v1}}. To make a fair comparison of five pre-trained monolingual S4-based language models for Vietnamese mentioned in the previous subsection (\ref{subsec:architecture}), after doing pre-processing phase with word segmentation or without word segmentation, we apply the BPE method \cite{sennrich-etal-2016-neural} to segment the sentences from the pre-training dataset with subword units, using a vocabulary of 64K subword types.

\subsection{Optimization}

We inherit the implementation\footnote{\url{https://github.com/HazyResearch/state-spaces}} of S4-based language model from the study of \citet{s4}. We set a maximum length of 512 subword tokens when training and evaluating five pre-trained monolingual S4-based language models for Vietnamese. Following \citet{s4}, we optimize models using AdamW \cite{DBLP:journals/corr/abs-1711-05101} with a single cosine learning rate cycle with a maximum of 40 epochs and a number of warmup steps is 10000. The initial learning rate was set to 0.0005. We use a batch size of 128 with gradient accumulation steps of 16 on a V100 GPU (16 GB). The training is performed on the Google Colaboratory\footnote{\url{https://colab.research.google.com/}}. Lastly, we trained each S4-based language model for 10 days, except 14 days for the model without word segmentation in pre-processing phase.

\section{Experimental Setup}

We evaluate the performance with or without the word segmentation phase of two traditional classifiers (Naive Bayes and Support Vector Machines) and one deep learning classifier (S4-based language modeling), on two well-known Vietnamese sentiment corpora, including the VLSP2016-SA corpus of technical article reviews from the news and social media and the UIT-VSFC corpus of the educational survey. In this paper, we note again that we focus on finding the effect of applying or not applying the word segmentation phase in the pre-processing data phase on deep learning models' Vietnamese sentiment classification performances. This means we do not focus on finding new state-of-the-art results.

\subsection{Corpora}

\begin{table}[H]
\centering
\caption{Number Of Samples Of Each Class In VLSP2016-SA Corpus}
\label{tab:stat:vlsp2016}
\resizebox{0.85\columnwidth}{!}{%
\begin{tabular}{|l|r|r|r|r|}
\hline
 & \multicolumn{1}{l|}{Negative} & \multicolumn{1}{l|}{Neutral} & \multicolumn{1}{l|}{Positive} & \multicolumn{1}{l|}{Overall} \\ \hline
\textbf{Training} & 1,526 & 1,519 & 1,530 & 4,575 \\ \hline
\textbf{Validation} & 174 & 181 & 170 & 525 \\ \hline
\textbf{Test} & 350 & 350 & 350 & 1,050 \\ \hline
\end{tabular}%
}
\end{table}

Table~\ref{tab:stat:vlsp2016} presents the statistics of the VLSP2016-SA \cite{vlsp2016sa} corpus. The original corpus does not have a validation set. Therefore, we have split the original training set into the new training and validation sets.

\begin{table}[H]
\centering
\caption{Number Of Samples Of Each Class In UIT-VSFC Corpus}
\label{tab:stat:uitvsfc}
\resizebox{0.85\columnwidth}{!}{%
\begin{tabular}{|l|r|r|r|r|}
\hline
 & \multicolumn{1}{l|}{Negative} & \multicolumn{1}{l|}{Neutral} & \multicolumn{1}{l|}{Positive} & \multicolumn{1}{l|}{Overall} \\ \hline
\textbf{Training} & 5,325 & 458 & 5,643 & 11,426 \\ \hline
\textbf{Validation} & 705 & 73 & 805 & 1,583 \\ \hline
\textbf{Test} & 1,409 & 167 & 1,590 & 3,166 \\ \hline
\end{tabular}%
}
\end{table}

Table~\ref{tab:stat:uitvsfc} presents the statistics of the UIT-VSFC \cite{vsfc} corpus. We followed the splitting of training, validation, and test sets by \citet{vsfc}.

\subsection{Training}
For traditional classifiers, we used the combination of uni-grams, bi-grams, and tri-grams as the features for the classification problem. We used the \texttt{scikit-learn} library \cite{scikit-learn} for implementing\footnote{\url{https://scikit-learn.org/stable/}} the Naive Bayes and Support Vector Machines algorithms. For each corpus, we did not do word segmentation and did word segmentation using one of \texttt{RDRsegmenter}, \texttt{uitnlp}, \texttt{pyvi}, or \texttt{underthesea} toolkits. The experiment result is different for Support Vector Machines if we used different random states. Therefore, we report the average results of 100 runs with 100 random states for fair comparisons.

For modern classifiers, we fine-tuned five pre-trained monolingual S4-based language models for Vietnamese without word segmentation with appropriate word segmentation toolkits (\texttt{RDRsegmenter}, \texttt{uitnlp}, \texttt{pyvi}, or \texttt{underthesea}). We used the max-pooling vector of all contextual representations at the last layer of the S4-based language model for all positions of the input sentence as the feature vector for classification. We optimize models using AdamW \cite{DBLP:journals/corr/abs-1711-05101} with a single cosine learning rate cycle with a maximum of 5 epochs for the UIT-VSFC corpus and 10 epochs for the VLSP2016-SA corpus. The warmup proportion is 0.025. We set a maximum length of 8192 subwords for an input text and a batch size of 32 when fine-tuning both UIT-VSFC and VLSP2016-SA corpora. The initial learning rate is 0.001, and the weight decay of AdamW is 0.01. Notably, because we did not tune hyper-parameters in our experiments, we report the average results of 100 runs with 100 random states for fair comparisons. Finally, we have done fine-tuning experiments for 4 days.

\section{Experimental Results}

\subsection{Main Results}

\begin{table}[H]
\centering
\caption{Average Evaluation For Classification Models Over 100 Runs On Test Set Of VLSP2016-SA Corpus (\%)}
\label{tab:vlsp2016-sa-result}
\resizebox{\columnwidth}{!}{%
\begin{tabular}{|c|l|ccc|}
\hline
\multicolumn{1}{|l|}{\multirow{2}{*}{\textbf{Model}}} & \multicolumn{1}{c|}{\multirow{2}{*}{\textbf{\begin{tabular}[c]{@{}c@{}}Word\\ Segmenter\end{tabular}}}} & \multicolumn{3}{c|}{\textbf{VLSP2016-SA}} \\ \cline{3-5} 
\multicolumn{1}{|l|}{} & \multicolumn{1}{c|}{} & \multicolumn{1}{c|}{\textbf{micro-$F_1$}} & \multicolumn{1}{c|}{\textbf{macro-$F_1$}} & \textbf{weighted-$F_1$} \\ \hline
\multirow{5}{*}{NB} & \texttt{None} & \multicolumn{1}{c|}{\textbf{67.24}} & \multicolumn{1}{c|}{\textbf{67.38}} & \textbf{67.38} \\ \cline{2-5} 
 & \texttt{RDRsegmenter} & \multicolumn{1}{c|}{66.38} & \multicolumn{1}{c|}{66.39} & 66.39 \\ \cline{2-5} 
 & \texttt{uitnlp} & \multicolumn{1}{c|}{66.00} & \multicolumn{1}{c|}{66.00} & 66.00 \\ \cline{2-5} 
 & \texttt{pyvi} & \multicolumn{1}{c|}{67.14} & \multicolumn{1}{c|}{67.13} & 67.13 \\ \cline{2-5} 
 & \texttt{underthesea} & \multicolumn{1}{c|}{65.62} & \multicolumn{1}{c|}{65.62} & 65.62 \\ \hline
\multirow{5}{*}{SVMs} & \texttt{None} & \multicolumn{1}{c|}{\textbf{67.96}} & \multicolumn{1}{c|}{\textbf{67.99}} & \textbf{67.99} \\ \cline{2-5} 
 & \texttt{RDRsegmenter} & \multicolumn{1}{c|}{67.56} & \multicolumn{1}{c|}{67.56} & 67.56 \\ \cline{2-5} 
 & \texttt{uitnlp} & \multicolumn{1}{c|}{67.33} & \multicolumn{1}{c|}{67.33} & 67.33 \\ \cline{2-5} 
 & \texttt{pyvi} & \multicolumn{1}{c|}{67.93} & \multicolumn{1}{c|}{67.88} & 67.88 \\ \cline{2-5} 
 & \texttt{underthesea} & \multicolumn{1}{c|}{66.77} & \multicolumn{1}{c|}{66.78} & 66.78 \\ \hline
\multirow{5}{*}{S4} & \texttt{None} & \multicolumn{1}{c|}{67.44} & \multicolumn{1}{c|}{67.42} & 67.42 \\ \cline{2-5} 
 & \texttt{RDRsegmenter} & \multicolumn{1}{c|}{68.14} & \multicolumn{1}{c|}{68.07} & 68.07 \\ \cline{2-5} 
 & \texttt{uitnlp} & \multicolumn{1}{c|}{\textbf{68.51}} & \multicolumn{1}{c|}{\textbf{68.47}} & \textbf{68.47} \\ \cline{2-5} 
 & \texttt{pyvi} & \multicolumn{1}{c|}{67.88} & \multicolumn{1}{c|}{67.83} & 67.83 \\ \cline{2-5} 
 & \texttt{underthesea} & \multicolumn{1}{c|}{67.85} & \multicolumn{1}{c|}{67.85} & 67.85 \\ \hline
\end{tabular}
}
\end{table}

Table~\ref{tab:vlsp2016-sa-result} presents average evaluation on the VLSP2016-SA corpus. Firstly, the Naive Bayes classifier trained without the word segmentation phase achieved higher performance than other Naive Bayes classifiers trained with the word segmentation phase. This also happened for the Support Vector Machines classifier. Secondly, S4-based classifiers trained with the word segmentation phase achieved higher performance than the S4-based classifier trained without the word segmentation phase. Especially, the S4-based classifier with the \texttt{uitnlp} toolkit achieved the higher result compared with \texttt{RDRsegmenter}, \texttt{pyvi}, and \texttt{underthesea}.

\begin{table}[H]
\centering
\caption{Average Evaluation For Classification Models Over 100 Runs On Test Set Of UIT-VSFC Corpus (\%)}
\label{tab:uit-vsfc-result}
\resizebox{\columnwidth}{!}{%
\begin{tabular}{|c|l|ccc|}
\hline
\multicolumn{1}{|l|}{\multirow{2}{*}{\textbf{Model}}} & \multicolumn{1}{c|}{\multirow{2}{*}{\textbf{\begin{tabular}[c]{@{}c@{}}Word\\ Segmenter\end{tabular}}}} & \multicolumn{3}{c|}{\textbf{UIT-VSFC}} \\ \cline{3-5} 
\multicolumn{1}{|l|}{} & \multicolumn{1}{c|}{} & \multicolumn{1}{c|}{\textbf{micro-$F_1$}} & \multicolumn{1}{c|}{\textbf{macro-$F_1$}} & \textbf{weighted-$F_1$} \\ \hline
\multirow{5}{*}{NB} & \texttt{None} & \multicolumn{1}{c|}{86.26} & \multicolumn{1}{c|}{59.83} & 84.07 \\ \cline{2-5} 
 & \texttt{RDRsegmenter} & \multicolumn{1}{c|}{86.99} & \multicolumn{1}{c|}{60.32} & 84.77 \\ \cline{2-5} 
 & \texttt{uitnlp} & \multicolumn{1}{c|}{87.11} & \multicolumn{1}{c|}{60.79} & 84.95 \\ \cline{2-5} 
 & \texttt{pyvi} & \multicolumn{1}{c|}{\textbf{87.18}} & \multicolumn{1}{c|}{\textbf{60.83}} & \textbf{84.99} \\ \cline{2-5} 
 & \texttt{underthesea} & \multicolumn{1}{c|}{86.99} & \multicolumn{1}{c|}{60.33} & 84.78 \\ \hline
\multirow{5}{*}{SVMs} & \texttt{None} & \multicolumn{1}{c|}{89.17} & \multicolumn{1}{c|}{73.48} & 88.55 \\ \cline{2-5} 
 & \texttt{RDRsegmenter} & \multicolumn{1}{c|}{\textbf{89.36}} & \multicolumn{1}{c|}{72.96} & 88.60 \\ \cline{2-5} 
 & \texttt{uitnlp} & \multicolumn{1}{c|}{89.35} & \multicolumn{1}{c|}{73.05} & \textbf{88.62} \\ \cline{2-5} 
 & \texttt{pyvi} & \multicolumn{1}{c|}{89.29} & \multicolumn{1}{c|}{\textbf{73.10}} & 88.54 \\ \cline{2-5} 
 & \texttt{underthesea} & \multicolumn{1}{c|}{89.27} & \multicolumn{1}{c|}{72.58} & 88.47 \\ \hline
\multirow{5}{*}{S4} & \texttt{None} & \multicolumn{1}{c|}{90.88} & \multicolumn{1}{c|}{75.82} & 90.11 \\ \cline{2-5} 
 & \texttt{RDRsegmenter} & \multicolumn{1}{c|}{\textbf{91.62}} & \multicolumn{1}{c|}{\textbf{77.84}} & \textbf{90.99} \\ \cline{2-5} 
 & \texttt{uitnlp} & \multicolumn{1}{c|}{91.36} & \multicolumn{1}{c|}{77.17} & 90.68 \\ \cline{2-5} 
 & \texttt{pyvi} & \multicolumn{1}{c|}{91.40} & \multicolumn{1}{c|}{77.74} & 90.80 \\ \cline{2-5} 
 & \texttt{underthesea} & \multicolumn{1}{c|}{91.35} & \multicolumn{1}{c|}{77.26} & 90.72 \\ \hline
\end{tabular}
}
\end{table}

Table~\ref{tab:uit-vsfc-result} presents average evaluation on the UIT-VSFC corpus. As we can see in Table~\ref{tab:uit-vsfc-result}, the Naive Bayes, Suppor Vector Machine, and S4-based classifier with the word segmentation phase achieved higher performance than other classifiers without the word segmentation phase. Especially, the S4-based classifier with the \texttt{RDRsegmenter} toolkit achieved the higher result compared with \texttt{uitnlp}, \texttt{pyvi}, and \texttt{underthesea}.

\subsection{Discussion}

As we can see in Table~\ref{tab:vlsp2016-sa-result}, using word segmentation for traditional classifiers like Naive Bayes and Support Vector Machines maybe achieve a lower performance than classifiers without word segmentation for the Vietnamese sentiment classification task. According to the study of \citet{vund2018}, the \texttt{uitnlp} achieved the higher F-score of word segmentation than \texttt{RDRsegmenter} \cite{datnq1} on the Vietnamese Treebank corpus for word segmentation \cite{nguyen-etal-2009-building}. However, the VLSP2016-SA corpus of technical article reviews from the news and social media. Hence, the social corpus's word segmentation criteria may differ from the Vietnamese Treebank corpus for word segmentation \cite{nguyen-etal-2009-building}. Finally, the potential reason explaining the traditional classifier with \texttt{uitnlp} toolkit achieved lower performance than the classifier with \texttt{RDR} is that the \texttt{uitnlp} toolkit was trained with over-fitting on the Vietnamese Treebank corpus for word segmentation \cite{nguyen-etal-2009-building}, which is not good when \texttt{uitnlp} toolkit faces the social corpus. This suggests that building Vietnamese word segmentation and part-of-speech tagging for social media text, like the study of \citet{10.1007/978-3-319-46675-0_26} is worthy of respect and attention.

In such an analytical way, in Table~\ref{tab:uit-vsfc-result}, the Naive Bayes, Support Vector Machine, and S4-based classifier with the word segmentation phase achieved higher performance than other classifiers without the word segmentation phase. This can be explained by the word segmentation criteria in the educational survey domain being close to the Vietnamese Treebank corpus for word segmentation \cite{nguyen-etal-2009-building}. Because the UIT-VSFC corpus was normalized by humans \cite{vsfc}.

\section{Conclusion}

This paper attempts to answer whether word segmentation is necessary for Vietnamese sentiment classification. According to comprehensive experimental results on two corpora, including the VLSP2016-SA corpus of technical article reviews from the news and social media and the UIT-VSFC corpus of the educational survey, we have two suggestions. Firstly, using traditional classifiers like Naive Bayes or Support Vector Machines, word segmentation maybe not be necessary for the Vietnamese sentiment classification corpus, which comes from the social domain. Secondly, word segmentation is necessary for Vietnamese sentiment classification when word segmentation is used before using the BPE method and feeding into the deep learning model. In this way, the \texttt{RDRsegmenter} is the stable toolkit for word segmentation among the \texttt{uitnlp}, \texttt{pyvi}, and \texttt{underthesea} toolkits because the \texttt{RDRsegmenter} achieved the stable performance on sentiment classification task in our experiments and has fast inference speed \cite{datnq1}.

\section*{Acknowledgement}
This research was supported by The VNUHCM-University of Information Technology's Scientific Research Support Fund.

{
    \balance
    \footnotesize
    \bibliography{main.bib}
}

\end{document}